\pdfoutput=1

\documentclass[11pt]{article}

\usepackage[preprint]{coling}

\usepackage{times}
\usepackage{latexsym}
\usepackage{booktabs}

\usepackage[T1]{fontenc}

\usepackage[utf8]{inputenc}
\usepackage{footnote}  
\makesavenoteenv{tabular}  
\usepackage{microtype}
\usepackage{inconsolata}
\usepackage{amsmath}

\usepackage{graphicx}

\usepackage{kotex}
\usepackage{multirow}
\usepackage{makecell} 

\usepackage{adjustbox}
\usepackage{tablefootnote}  
\usepackage{inconsolata}
\usepackage{xcolor}
\usepackage{xspace}
\newcommand{\modelName}{\textsc{Pad}\xspace}

\title{\modelName: Towards Efficient Data Generation for Transfer Learning Using Phrase Alignment}

\author{
    Jong Myoung Kim$^{1,2}$ \hspace{0.3cm}
    Young-Jun Lee$^{2}$ \hspace{0.3cm}
    Ho-Jin Choi$^{2}$ \hspace{0.3cm}
    Sangkeun Jung$^{3}$ \hspace{0.3cm}
    \\
    $^1$SK-telecom \hspace{0.3cm}
    $^2$School of Computing, KAIST \hspace{0.3cm} \\
    $^3$The Division of Computer Convergence, Chugnam National University\\
    \href{mailto:jmkim71@sk.com}{\tt gray.apl@gmail.com} \hspace{0.3cm} 
    \{\href{mailto:yj2961@kaist.ac.kr}{\tt yj2961},\href{mailto:hojinc@kaist.ac.kr}{\tt hojinc}\}{\tt @kaist.ac.kr} \hspace{0.3cm} 
    \href{mailto:hugman@cnu.ac.kr}{\tt hugman@cnu.ac.kr}
}

\begin{document}
\maketitle
\begin{abstract}

Building training datasets for specific tasks is essential for service deployment, but it often requires significant time and cost. To address this issue, transfer learning using abundant English data has been widely adopted in non-English language model development. We propose a novel method, Phrase Aligned Data (\modelName), to further improve the efficiency of transfer learning. \modelName is created by applying phrase alignment, a core module of Statistical Machine Translation (SMT), to English data and leverages the syntactic flexibility of languages like Korean to maximize transfer learning effects. Our experiments show that \modelName not only enhances transfer learning performance but also significantly reduces the cost and time of data construction, demonstrating its value for both academic research and industrial applications.

\end{abstract}

\section{Introduction}

The rapid advancements in large-scale language models have spurred the development of various services. Many companies build proprietary datasets to fine-tune their models and enhance service quality, but constructing such high-quality datasets is costly and time-consuming.

Non-English-speaking NLP engineers often address this challenge by leveraging the abundant English resources available. For instance, the Korean KLUE~\citep{park2021klue} NLI dataset has only 24,000 entries, while the English GLUE~\citep{wang2018glue} dataset has about 300,000 entries. Despite Korean being a relatively resource-rich language, there is still a significant gap in the size and diversity of data between English and non-English languages~\citep{wang2022super, joshi2020state}. As a result, English resources are widely used to enhance non-English models through cross-lingual transfer learning, providing a cost-effective and time-efficient solution.

To improve the efficiency of using English data for Korean models, we focus on two key aspects: (1) \textbf{Expression in the Target Language}: Models perform better when the data is expressed in the target language, even if it conveys the same meaning~\citep{imani2023glot500}. This suggests that translating English data into Korean before fine-tuning can improve performance; (2) \textbf{Syntactic Flexibility of Korean}: Korean uses case markers to indicate the roles of sentence components, allowing for flexible word order. This means that even data with an incomplete word order can still be useful as training data~\citep{JMKIM}.

Based on these insights, we propose \modelName, a low-cost method that converts English data into a format suitable for training Korean models. \modelName is generated using phrase alignment, a core component of SMT, which was widely used before the advent of Neural Machine Translation (NMT). Phrase alignment probabilistically aligns source language phrases with corresponding target language phrases. \modelName uses this mechanism to create datasets that, while potentially having incomplete word order, preserve the meaning of each phrase and are expressed in the target language.

Our experiments show that \modelName effectively transforms English data into a form suitable for Korean tasks and can be produced with minimal resources, time, and cost compared to direct data construction.

Our contributions are as follows:
\begin{itemize}
    \item We introduce \modelName, combining SMT’s phrase alignment technique with the syntactic characteristics of Korean.
    \item We demonstrate that \modelName effectively adapts English data for transfer learning in Korean.
    \item \modelName provides significant time and cost savings over direct dataset construction and serves as a baseline for new data creation, underscoring its industrial value.
\end{itemize}

\section{Related Works}
\subsection{Cross-Lingual Transfer Learning}
Cross-lingual transfer learning has been widely used to enhance non-English language models by leveraging pre-trained English models~\citep{raffel2020exploring}. This method utilizes the abundant English data to fine-tune models for low-resource languages. However, performance often falls short compared to models trained on native target language data. Constructing high-quality target language datasets is expensive~\citep{schuster2018cross}, making translated English data a practical alternative. Although translated data underperforms compared to native data, it significantly improves performance over using untranslated English data~\citep{gaim2023question}. Our approach harnesses the richness of English data to achieve results close to those obtained with translated data.

\subsection{Word Order}
Understanding word order is essential for cross-lingual tasks, as different languages exhibit distinct syntactic structures. Prior research has shown that language models can benefit from training on corpora with disrupted word orders, enhancing their robustness~\citep{sinha2021masked}. Korean, with its flexible word order, allows for training on sentences with incomplete word structures~\citep{JMKIM}. This flexibility can be leveraged to extend the training dataset and improve model performance.

\subsection{Statistical Machine Translation (SMT)} \label{ch:SMT}
SMT uses statistical models for translation and is composed of alignment and reordering stages~\citep{koehn2009statistical}. During alignment, source language phrases are mapped to their corresponding target language phrases using parallel corpora. Reordering adjusts the sentence structure to conform to the target language’s natural syntax, addressing structural differences like English (S-V-O) and Korean (S-O-V).

Reordering remains a major challenge in SMT, as language pairs with different word orders often result in unnatural translations~\citep{koehn2009statistical}. While models such as distortion models attempt to resolve this, they struggle to handle complex reordering, affecting the overall fluency and coherence of the translated text.

\begin{table*}[ht]
\begin{adjustbox}{width=\linewidth}
\begin{tabular}{c|c|c}
\hline
Data   type       & Word   order & Example                                                    \\ \hline
English           & Complete     & I once met a physicists who held the strong AI position. \\
\modelName               & Incomplete   & 나는\textsubscript{(I)} 한 번\textsubscript{(once)} 만났다\textsubscript{(met)} 물리학자를\textsubscript{(a physicists)} 지지하는\textsubscript{(who held)} 강한 인공지능\textsubscript{(the strong AI)} 입장을\textsubscript{(position)}                          \\
Translated   Data & Complete     & 나는\textsubscript{(I)} 한 번\textsubscript{(once)} 강한 인공지능\textsubscript{(the strong AI)} 입장을\textsubscript{(position)} 지지하는\textsubscript{(who held)} 물리학자를\textsubscript{(a physicists)} 만났다\textsubscript{(met)}.                   \\ \hline
\end{tabular}
\end{adjustbox}

\caption{Examples of English data, \modelName data, and translation-generated data. Korean phrase meanings are shown in parentheses and superscripts for clarity (annotations are excluded from the actual data).}

\label{tab:dataExample}
\end{table*}
\section{Phrase Aligned Data: \modelName}
We propose a novel data processing method to maximize the performance of transfer learning while ensuring cost efficiency. Our approach is inspired by two key observations: (1) training data expressed in the target language tends to be more effective, even when it conveys the same meaning as the source language~\citep{imani2023glot500}, and (2) data with incomplete word order can still be valuable for training, especially for languages like Korean~\citep{JMKIM}. Based on these observations, we define \modelName, a data processing method that transforms English data into a format optimized for training Korean models.

\modelName is created by applying SMT procedures up to the phrase alignment stage, producing data that follows the word order of English but replaces each phrase with its corresponding Korean expression. We expect \modelName to achieve the following effects:

\begin{itemize}
    \item Show higher training efficiency compared to using raw English data.
    \item Demonstrate greater cost efficiency than constructing high-quality Korean datasets or manually translating English data.
    \item Improve stability and effectiveness by excluding the reorder stage, which is known to be a limitation of traditional SMT systems.
\end{itemize}

In this study, we used Moses~\citep{koehn2007moses}, the most widely known SMT system, to generate \modelName. Table~\ref{tab:dataExample} provides an example of original English data, \modelName data, and data generated through translation methods.

\subsection{Moses and its Training} 
Moses uses GIZA++ for phrase alignment based on the IBM models and performs reordering with a distortion model~\citep{casacuberta2007giza++}. In this experiment, we used IBM Model 4 and trained Moses on 1.7 million parallel sentences from the AI-hub\footnote{A data platform operated by the Korea National Information Society Agency (NIA)} corpus, which includes translated and verified data from sources such as news, government websites, legal documents, and various conversational contexts. The source English data was tokenized using the Moses tokenizer, and the target Korean data was tokenized at the word level, as the reordering stage was omitted.

\section{Experiments}
\subsection{Experiment Design}
We evaluated the effectiveness of \modelName as a data construction and processing method for transfer learning by: (1) conducting experiments on benchmarks available in both English and Korean, (2) constructing datasets of varying sizes using different data construction methods, and (3) fine-tuning models on these datasets to measure performance. Since \modelName aims to improve transfer learning using English data, we constructed datasets primarily from English data and evaluated performance on Korean test sets.

\subsection{Baselines - Data Construction Methods}
\subsubsection{Human-Generated Data}\label{sec:humangeneratedData}
\paragraph{English (English)}
Transfer learning with English data is the most basic alternative when target language data is not available. While this method incurs no additional construction cost, it tends to underperform compared to native target language data. This dataset was also used as the source for the machine-generated data described below.

\paragraph{Native Korean (Korean)}
Data created by native speakers of the target language is considered the highest quality training data. However, this method incurs the highest time and cost. We used a native Korean benchmark dataset for this experiment.

\subsubsection{Machine-Generated Data}\label{sec:MGD}
Translating English data into the target language is a common approach to improving transfer learning efficiency. High-quality translated data is known to perform between native data and English data~\citep{imani2023glot500}. We used the English dataset from Section \ref{sec:humangeneratedData} to create the following machine-generated datasets.

\paragraph{High-Quality NMT (NMT\texorpdfstring{\textsubscript{GPT-4}}{NMTGPT-4})}
We used GPT-4 turbo to construct the translated dataset. Since GPT-4 is considered one of the highest-performing machine translation models currently available~\citep{jiao2023chatgpt}, it serves as a realistic performance benchmark for \modelName. The cost for this method was primarily due to API usage fees.

\paragraph{Local-Tuned NMT (NMT\texorpdfstring{\textsubscript{mT5}}{NMTmT5},NMT\texorpdfstring{\textsubscript{mGPT}}{NMTmGPT})}
\modelName requires parallel corpora for its creation. To test \modelName’s performance against locally-tuned models, we used NMT models trained on the same parallel corpus. These models incur hardware and time costs for tuning. We used two backbone models: mT5-large and mGPT, as detailed in Section \ref{ch:backbone}.

\paragraph{SMT (SMT\texorpdfstring{\textsubscript{Moses}}{SMTMoses})}
We excluded the `reorder' stage of the SMT process to improve stability and efficiency. To verify this, we trained Moses on the parallel corpus used for \modelName generation and performed the full SMT process to create the translated data. As this method reconstructs phrases into complete sentences, we tokenized the Korean data using KoNLPy~\citep{park2014konlpy} and applied 3-gram kenLM with `msd-bidirectional-fe` reordering.

\begin{table}[t]
\begin{adjustbox}{width=\linewidth}
\begin{tabular}{c||cc||cc||c}
\hline
Language & \multicolumn{2}{c||}{Korean}                               & \multicolumn{2}{c||}{English}                               & \multirow{2}{*}{Criterion} \\ \cline{1-5} 
Task     & \multicolumn{1}{c|}{Publisher}             & Dataset name & \multicolumn{1}{c|}{Publisher}             & Dataset name &                          \\ \hline
SA       & \multicolumn{1}{c|}{\makecell{NSMC\\\citep{Park:2016}}}                  & SA           & \multicolumn{1}{c|}{\multirow{3}{*}{\makecell{GLUE\\\citep{wang2018glue}}}} & SA           & F1 score              \\ \cline{1-3} \cline{5-6} 
NLI      & \multicolumn{1}{c|}{\multirow{2}{*}{\makecell{KLUE\\\citep{park2021klue}}}} & NLI          & \multicolumn{1}{c|}{}                      & MNLI         & Accuracy              \\ \cline{1-1} \cline{3-3} \cline{5-6} 
STS      & \multicolumn{1}{c|}{}                      & STS          & \multicolumn{1}{c|}{}                      & STS          & Pearson's r              \\ \hline
\end{tabular}

\end{adjustbox}
\caption{Basic information of the benchmark datasets used in the experiments.} \label{tab:datasetInfo}
\end{table}

\subsection{Benchmark Tasks}
We tested the utility of aligned data through three tasks. We combined the original datasets' labeled training and validation sets to create our train, validation, and test datasets. The test set contained 2000 instances, and we incrementally increased the training data by 1000 instances for each experiment. The training and validation data were split in a 9:1 ratio. Basic information about the benchmark datasets used in this experiment is provided in Table~\ref{tab:datasetInfo}, and detailed information can be found in Appendix~\ref{ap:benchmarkInfo}.

\paragraph{Sentiment Analysis (SA)} SA detects and classifies the emotional tone of text, categorizing it into positive or negative sentiments. It is applied in movie reviews. Models predict sentiment based on linguistic features and contextual understanding.

\paragraph{Natural Language Inference (NLI)} NLI determines the logical relationship between two sentences: a premise and a hypothesis, classifying the relationship into entailment, contradiction, or neutral.

\paragraph{Semantic Textual Similarity (STS)} STS measures the degree of semantic equivalence between two text segments, providing a similarity score. It is used in information retrieval and clustering.

\begin{table}[t]
\begin{adjustbox}{width=\linewidth}
\begin{tabular}{c|c|c|c|c}
\hline
                              & Architecture & Publisher  & Model name                                                                 & Parameter Size \\ \hline
\multirow{2}{*}{Multilingual} & GPT-3        & AI-forever & \makecell{mGPT \\ \citep{https://doi.org/10.48550/arxiv.2204.07580}}       & 580M           \\ \cline{2-5} 
                              & T5           & Google     & \makecell{mT5-base \\ \citep{xue-etal-2021-mt5}}                           & 1.3B           \\ \hline
\multirow{2}{*}{Monolingual}  & GPT-3        & SK telecom & GPT-trinity\tablefootnote{https://huggingface.co/skt/ko-gpt-trinity-1.2B-v0.5}  & 1.2B           \\ \cline{2-5} 
                              & T5           & PAUST      & pko-T5-base\tablefootnote{https://huggingface.co/paust/pko-t5-base}             & 220M           \\ \hline
\end{tabular}

\end{adjustbox}
\caption{Information of the models used in the experiments.} \label{tab:modelInfo}
\end{table}
\subsection{Models}\label{ch:backbone}
We used two architectures, T5 and GPT-3, and prepared both multilingual and monolingual language models for our experiments. The structure and characteristics of each model are summarized below, with additional details in Table~\ref{tab:modelInfo} and Appendix~\ref{ap:modelDetail}.

\paragraph{\texttt{T5 Series}}
The T5 (Text-to-Text Transfer Transformer) model by Google~\citep{xue2020mt5} unifies all NLP tasks into a text-to-text format, making it highly versatile. Tasks like translation, summarization, and question answering are handled by converting text inputs into outputs, enabling effective transfer learning across various NLP tasks.

\paragraph{\texttt{GPT3 Series}}
GPT-based models have shown strong performance across many NLP tasks. Due to hardware constraints, we selected a smaller GPT-3 variant suitable for our environment.

\subsection{Implementation Details}
We conducted fine-tuning and generation experiments for syntactic flexibility using an NVIDIA A100 40GB GPU. 
We employed the Adam optimizer with decoupled weight decay~\citep{ loshchilov2017decoupled} and set a learning rate of 5e-5 with a warm-up period spanning 3 epochs.
Detailed information on the OS and library versions is provided in Appendix~\ref{ap:impDetail}.

\begin{table*}[t]
\begin{adjustbox}{width=\linewidth}
\begin{tabular}{cccccccccccccc}
\hline
\multicolumn{1}{c|}{Task}      & \multicolumn{5}{c|}{SA (F1 score)}                                    & \multicolumn{5}{c|}{NLI (Accuracy)}                                   & \multicolumn{3}{c}{STS (Pearson's r)} \\ \hline
\multicolumn{1}{c|}{Data Size} & 4K    & 8K    & 12K   & 16K   & \multicolumn{1}{c|}{20K}   & 4K    & 8K    & 12K   & 16K   & \multicolumn{1}{c|}{20K}   & 3K     & 5K     & 7K    \\ \hline
\multicolumn{1}{l|}{English}   & 0.696 & 0.716 & 0.730 & 0.729 & \multicolumn{1}{c|}{0.727} & 0.269 & 0.307 & 0.417 & 0.474 & \multicolumn{1}{c|}{0.572} & 0.694  & 0.665  & 0.819 \\
\multicolumn{1}{l|}{Korean}    & \cellcolor[HTML]{9B9B9B}0.762 & 0.822 & 0.830 & 0.817 & \multicolumn{1}{c|}{0.826} & \cellcolor[HTML]{9B9B9B}0.384 & \cellcolor[HTML]{9B9B9B}0.565 & 0.686 & 0.667 & \multicolumn{1}{c|}{0.724} & \cellcolor[HTML]{9B9B9B}0.779  & \cellcolor[HTML]{9B9B9B}0.744  & 0.880 \\
\multicolumn{1}{l|}{NMT\textsubscript{GPT-4}}     & \cellcolor[HTML]{9B9B9B}0.748 & \cellcolor[HTML]{9B9B9B}0.755 & \cellcolor[HTML]{9B9B9B}0.765 & \cellcolor[HTML]{9B9B9B}0.769 & \multicolumn{1}{c|}{0.772} & \cellcolor[HTML]{9B9B9B}0.389 & \cellcolor[HTML]{9B9B9B}0.569 & \cellcolor[HTML]{9B9B9B}0.672 & \cellcolor[HTML]{9B9B9B}0.666 & \multicolumn{1}{c|}{0.711} & \cellcolor[HTML]{9B9B9B}0.752  & \cellcolor[HTML]{9B9B9B}0.691  & 0.851 \\
\multicolumn{1}{l|}{NMT\textsubscript{mGPT}}    & 0.705 & 0.725 & 0.755 & 0.750  & \multicolumn{1}{l|}{0.756} & 0.329 & 0.410  & 0.514 & 0.578 & \multicolumn{1}{c|}{0.61}  & 0.745  & 0.671  & 0.841 \\
\multicolumn{1}{l|}{NMT\textsubscript{mT5}}     & 0.719 & 0.728 & 0.736 & 0.747 & \multicolumn{1}{c|}{0.757} & 0.407 & 0.497 & 0.593 & 0.670  & \multicolumn{1}{c|}{0.679} & 0.762  & 0.704  & 0.840  \\
\multicolumn{1}{l|}{SMT\textsubscript{Moses}}     & 0.735 & 0.750  & 0.749 & 0.753 & \multicolumn{1}{c|}{0.742} & 0.360  & 0.495 & 0.555 & 0.612 & \multicolumn{1}{c|}{0.663} & 0.736  & 0.671  & 0.821 \\
\multicolumn{1}{l|}{\modelName}       & 0.735 & 0.737 & 0.752 & 0.762 & \multicolumn{1}{c|}{\textbf{0.770}}  & 0.352 & 0.509 & 0.639 & 0.665 & \multicolumn{1}{c|}{\textbf{0.681}} & 0.735  & 0.677  & \textbf{0.844} \\ \hline
\multicolumn{14}{c}{Performance   Evaluation with mGPT (Multilingual Model)}                                                                                                       \\ \midrule
\multicolumn{1}{l|}{Korean}    & 0.871 & 0.893 & 0.892 & 0.895 & \multicolumn{1}{c|}{0.902} & 0.777 & 0.779 & 0.811 & 0.816 & \multicolumn{1}{c|}{0.834} & 0.834  & 0.842  & 0.844 \\
\multicolumn{1}{l|}{NMT\textsubscript{GPT-4}}     & \cellcolor[HTML]{9B9B9B}0.839 & 0.851 & 0.859 & 0.863 & \multicolumn{1}{c|}{0.866} & \cellcolor[HTML]{9B9B9B}0.731 & \cellcolor[HTML]{9B9B9B}0.761 & 0.773 & 0.766 & \multicolumn{1}{c|}{0.782} & \cellcolor[HTML]{9B9B9B}0.824  & 0.830   & 0.831 \\
\multicolumn{1}{l|}{NMT\textsubscript{mGPT}}    & 0.809 & 0.816 & 0.814 & 0.831 & \multicolumn{1}{c|}{0.847} & 0.663 & 0.715 & 0.747 & 0.738 & \multicolumn{1}{c|}{0.758} & 0.776  & 0.789  & 0.819 \\
\multicolumn{1}{l|}{NMT\textsubscript{mT5}}     & 0.818 & 0.823 & 0.841 & 0.835 & \multicolumn{1}{c|}{0.86}  & 0.669 & 0.707 & 0.747 & 0.750 & \multicolumn{1}{c|}{0.753} & 0.815  & 0.811  & 0.816 \\
\multicolumn{1}{l|}{SMT\textsubscript{Moses}}     & 0.827 & 0.843 & 0.845 & 0.855 & \multicolumn{1}{c|}{0.856} & 0.723 & 0.755 & 0.764 & 0.757 & \multicolumn{1}{c|}{0.76}  & 0.816  & 0.828  & 0.824 \\
\multicolumn{1}{l|}{\modelName}       & 0.832 & 0.833 & 0.845 & 0.845 & \multicolumn{1}{c|}{\textbf{0.849}} & 0.715 & 0.742 & 0.746 & \textbf{0.762} & \multicolumn{1}{c|}{0.754} & 0.818  & 0.826  & \textbf{0.829} \\ \hline
\multicolumn{14}{c}{Performance   Evaluation with GPT-Trinity (Monolingual Model)}                                                                                                 \\ \midrule
\multicolumn{1}{l|}{Korean}    & 0.856 & 0.879 & 0.885 & 0.895 & \multicolumn{1}{c|}{0.902} & \cellcolor[HTML]{9B9B9B}0.708 & \cellcolor[HTML]{9B9B9B}0.732 & 0.779 & 0.794 & \multicolumn{1}{c|}{0.803} & 0.868  & 0.873  & 0.875 \\
\multicolumn{1}{l|}{NMT\textsubscript{GPT-4}}     & \cellcolor[HTML]{9B9B9B}0.827 & 0.842 & 0.849 & 0.855 & \multicolumn{1}{c|}{0.863} & \cellcolor[HTML]{9B9B9B}0.688 & \cellcolor[HTML]{9B9B9B}0.701 & \cellcolor[HTML]{9B9B9B}0.731 & 0.754 & \multicolumn{1}{c|}{0.75}  & \cellcolor[HTML]{9B9B9B}0.853  & 0.858  & 0.863 \\
\multicolumn{1}{l|}{NMT\textsubscript{mGPT}}    & 0.777 & 0.811 & 0.824 & 0.827 & \multicolumn{1}{c|}{0.839} & 0.579 & 0.657 & 0.682 & 0.711 & \multicolumn{1}{c|}{0.723} & 0.754  & 0.811  & 0.844 \\
\multicolumn{1}{l|}{NMT\textsubscript{mT5}}     & 0.780 & 0.829 & 0.845 & 0.852 & \multicolumn{1}{c|}{0.852} & 0.579 & 0.695 & 0.719 & 0.746 & \multicolumn{1}{c|}{0.734} & 0.795  & 0.836  & 0.817 \\
\multicolumn{1}{l|}{SMT\textsubscript{Moses}}     & 0.804 & 0.827 & 0.829 & 0.831 & \multicolumn{1}{c|}{0.839} & 0.6   & 0.711 & 0.732 & 0.730 & \multicolumn{1}{c|}{0.726} & 0.843  & 0.851  & 0.852 \\
\multicolumn{1}{l|}{\modelName}       & 0.795 & 0.821 & 0.830 & 0.832 & \multicolumn{1}{c|}{\textbf{0.841}} & 0.597 & 0.706 & 0.726 & 0.728 & \multicolumn{1}{c|}{\textbf{0.732}} & 0.835  & 0.854  & \textbf{0.855} \\ \hline
\multicolumn{14}{c}{Performance   Evaluation with PKO-T5 (Monolingual Model)}                                                                                                     
\end{tabular}
\end{adjustbox}
\caption{The performance evaluation results compare the \modelName model with various baselines across three tasks: SA, NLI, and STS. Among the models trained with \modelName data, the \textbf{best-performing case} is highlighted in bold. Additionally, results for the Korean and NMT\textsubscript{GPT-4} models are shaded in \textcolor[HTML]{9B9B9B}{gray} if their performance is lower than the best \modelName model performance. For instance, in the Sentiment Analysis task using the mGPT model, the \modelName model trained with 20K data outperforms the NMT\textsubscript{GPT-4} model trained with 16K data.} \label{tab:mainResult}

\end{table*}
\begin{table}[h]
\begin{adjustbox}{width=\linewidth}
\begin{tabular}{l|lll|lll}
\hline
          & \multicolumn{3}{c|}{SA} & \multicolumn{3}{c}{NLI} \\ \hline
          & 30K    & 40K    & 50K   & 30K    & 40K    & 50K   \\ \hline
mGPT-\modelName & 0.759  & 0.764  & 0.770  & 0.687  & 0.693  & 0.701 \\
GPT Trinity-\modelName & 0.860   & 0.853  & 0.862 & 0.763  & 0.754  & 0.767 \\
PKO-T5-\modelName & 0.843  & 0.846  & 0.845 & 0.739  & 0.741  & 0.741 \\ \hline
\end{tabular}
\end{adjustbox}
\caption{Results of experiments leveraging the abundance of English data. Performance improvements were achieved without incurring additional costs.}\label{tab:overFarm}
\end{table}
\section{Experimental Results}

Table~\ref{tab:mainResult} shows the results of three benchmark tests conducted using three different models.  
(The mT5 model was also evaluated, but due to training issues, it only produced a single label and value. More details are provided in Appendix~\ref{ap:mt5Results}.)  
The rows in the table represent the data construction methods used, while the columns indicate the size of the generated and trained data.  
Although data construction was performed in 1K increments, the table is presented in 4K increments (2K for STS) for better readability.  
More detailed experimental results are provided in the Appendix~\ref{ap:detailedResults}.

\paragraph{vs. English}  
We compared the performance of models trained on English data using the mGPT model.  
In all cases, the models trained on \modelName data outperformed those trained on English data.

\paragraph{vs. Korean \& NMT\textsubscript{GPT-4}}  
The Korean and NMT\textsubscript{GPT-4} datasets serve as high-quality reference data in the Korean data construction tasks.  
When trained with the same amount of data, the models trained on \modelName data did not surpass those trained on these two datasets.  
However, despite being constructed using only publicly available resources and CPUs, the models trained on \modelName data still showed performance similar to those trained on the reference datasets.  
In Table~\ref{tab:mainResult}, the highest performance cases of the models trained on \modelName data are highlighted in bold, and the results of the Korean and NMT\textsubscript{GPT-4} models that showed lower performance than the best \modelName results are shaded in gray.  
For example, in the Sentiment Analysis task with the mGPT model, the model trained on 20K \modelName data outperformed the model trained on 16K NMT\textsubscript{GPT-4} data.

\paragraph{vs. Local-Tuned NMT (NMT\textsubscript{mT5}, NMT\textsubscript{mGPT})}  
This section shows the results of using the parallel corpus used for \modelName construction to train NMT models.  
Due to the limitations of the A100 40GB GPU, the range of models was restricted to mT5 and mGPT.  
Additionally, under the constraint of using a 1.7M parallel corpus, the models trained on \modelName data showed 3.9\% higher performance than NMT\textsubscript{mGPT} and 0.6\% higher performance than NMT\textsubscript{mT5}.  
This difference was calculated as the average of the relative performance differences, using the formula: \(\text{average} \left( \frac{\text{score}(\modelName) - \text{score}(NMT)}{\text{score}(\modelName)} \right)\).
Furthermore, training for 5 epochs required over 70 hours, highlighting the high time cost of NMT training.

\paragraph{vs. SMT\textsubscript{Moses}}  
Overall, the models trained on \modelName data showed performance similar to SMT\textsubscript{Moses}, with a difference within 0.5\%.  
This result indicates that the impact of the two data generation methods on Korean language training is similar.

\subsection{Utilizing the Abundance of English Data}
As mentioned in Section 1, English language resources show a significant difference in quantity and diversity compared to other languages.  
By leveraging the abundance of English data, we were able to generate additional PAD data without incurring additional costs.  
Table~\ref{tab:overFarm} shows the performance of models trained on the additional PAD data.  
Although the performance did not surpass that of NMT\textsubscript{GPT-4} 20K, it was able to achieve a closer performance, indicating that the additional PAD data contributed to improving the model's performance.

\begin{figure}[ht]
    \centering
    \includegraphics[width=0.8\linewidth]{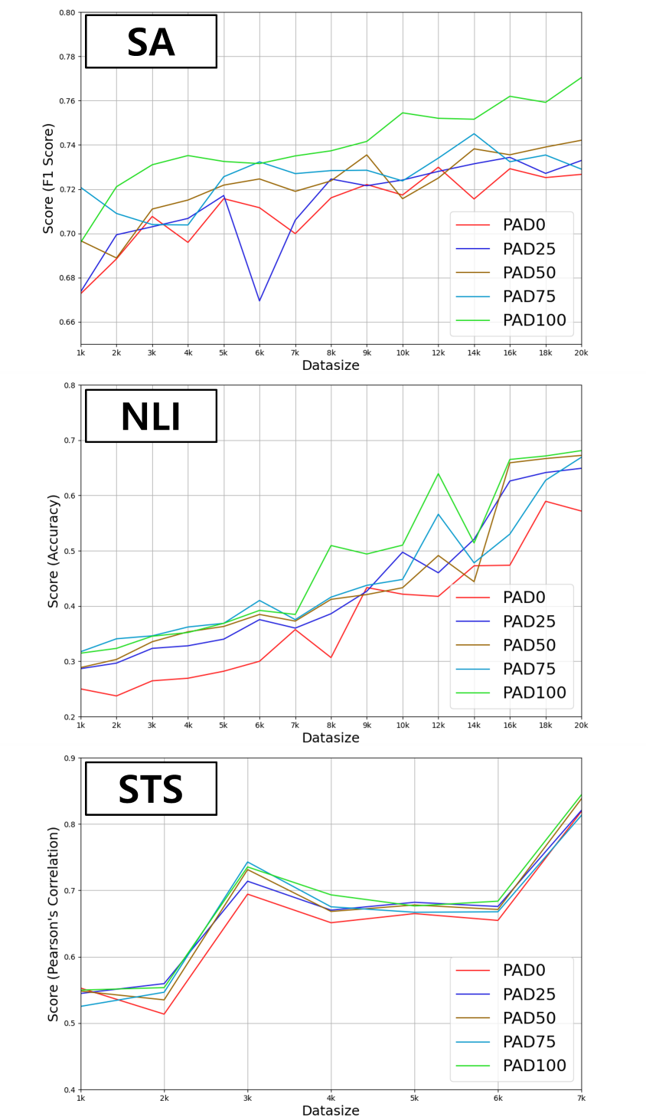} 
    \caption{Benchmark performance comparison by Korean-English mixed sentence ratios.}
    \label{fig:additionalWhole}
\end{figure}

\subsection{Mixed Korean-English Data}

While examining the generated data, we identified mixed Korean-English sentences due to SMT’s policy of outputting the source language expression when encountering unknown inputs. (Figure~\ref{fig:mixedSMTexample} in Appendix~\ref{sec:support} shows how SMT handles such expressions.) We investigated if the language model could learn from mixed Korean-English sentences and its impact on performance.

We generated mixed Korean-English sentences at various ratios (0\%, 25\%, 50\%, 75\%, and 100\%) and trained models with these datasets. SMT produced mixed language sentences due to a lack of training data, but we did not create mixed sentences by reducing training data. This is because the reduction in training data does not directly reflect the language ratio in the mixed sentences. Instead, we analyzed the logs from \modelName creation to determine the probability of phrase translation, generating data with stable ratios of mixed sentences.

Figure~\ref{fig:additionalWhole} shows the results. Models trained with mixed sentences outperformed those trained with only English data but did not match the performance of models trained with fully translated data. Models with a higher proportion of Korean (\modelName100) performed well due to the test set being in Korean, demonstrating better controllability compared to LLMs that output nonsense or repetitive tokens when undertrained. (Figure~\ref{fig:looping} in Appendix shows the looping phenomenon observed in translation data construction using mT5.)

\section{Discussion}
\modelName data demonstrated its effectiveness as a data processing method for transfer learning by outperforming models trained on English data.  
In addition, \modelName surpassed the performance of local-tuned NMT and SMT models but did not exceed the performance of models trained on the highest-quality Korean data or translation data (NMT\textsubscript{GPT-4}).  
Nevertheless, \modelName provides great value in terms of utility, as it incurs minimal costs, such as GPU requirements or API fees, and can achieve performance levels close to that of the NMT\textsubscript{GPT-4} model when combined with the abundance of English data.

In situations where \modelName training is insufficient, it has been observed that the output contains a mixture of Korean and English, rather than unexpected outputs from the LLM. This controllable behavior serves as an advantage.  
The occurrence of mixed-language outputs suggests that the model can be controlled to operate stably even when the training data is insufficient.

\modelName offers significant advantages in terms of cost, time, and computational efficiency, and can also provide an approximate lower bound for performance when generating high-quality data, thereby demonstrating its industrial value.

However, this study was conducted only in Korean, which limits its generalizability to other languages.  
Additionally, we did not confirm the correlation between complex sentence structures and the effectiveness of \modelName. Future research will explore the applicability of \modelName to various languages and sentence structures.

\section{Applications}
We are applying this research in the following ways in real-world scenarios: 
\begin{itemize}
    \item During the planning phase of a service, where there is a lack of training data or budget, \modelName is used as a method for generating demo data.  
    \item In the early stages of service development, when there is insufficient training data, \modelName is utilized as supplemental data to augment the existing training set. As internally generated training data becomes sufficient, the use of \modelName data gradually fades out.
\end{itemize}

\section{Conclusion}
We propose \modelName by combining the benefits of native language training data, Korean syntactic flexibility, and abundant English data. \modelName reliably improves model performance, is cost-efficient, and offers controllability, suggesting significant industrial value in scenarios where cost and stability are critical. Future research should explore the applicability of \modelName across various languages.

\bibliography{custom}

\appendix
\section{Benchmark Dataset Information}\label{ap:benchmarkInfo}
\section{Detailed-Model Informations}\label{ap:modelDetail}
\subsection{T5 models}
The T5 (Text-to-Text Transfer Transformer) model is a transformer-based architecture developed by Google. It converts all NLP tasks into a text-to-text format, making it highly versatile. Tasks such as translation, summarization, and question answering are approached by feeding the model text inputs and training it to generate the corresponding text outputs. This unified approach allows the T5 model to leverage transfer learning effectively across various NLP tasks.

\paragraph{mT5} is a multilingual version of the T5 model developed by Google~\citep{xue-etal-2021-mt5}, trained on multilingual corpora covering 101 languages. mT5 handles various NLP tasks using a unified text-to-text format and includes models with up to 13 billion parameters. It utilizes the SentencePiece tokenizer for processing multilingual text.

\paragraph{Paust T5} is a Korean-optimized T5 model fine-tuned on Korean corpora~\footnote{https://huggingface.co/paust/pko-t5-base}. The model comprises 220 million parameters and demonstrates excellent performance in Korean NLP tasks. It employs a SentencePiece tokenizer optimized for processing Korean text.
\subsection{GPT based models}
Recently, GPT-based models have demonstrated excellent performance across various NLP tasks.
Due to the limitations of our development environment, we selected a model that follows the GPT-3 architecture but is of a manageable size for our capabilities.
\paragraph{ai-forever mGPT} is a multilingual GPT model known for its strong performance across various NLP tasks~\citep{https://doi.org/10.48550/arxiv.2204.07580}. This model has 130 million parameters and is trained on multilingual corpora. It is designed to process text in multiple languages and follows GPT-3 architecture using GPT-2 source.

\paragraph{skt ko-gpt-trinity-1.2B-v0.5} is a Korean-optimized GPT model developed by SKT, featuring 1.2 billion parameters~\footnote{https://huggingface.co/skt/ko-gpt-trinity-1.2B-v0.5}. Trained on Korean corpora, this model excels in Korean NLP tasks. It uses the BPE tokenizer, which is optimized for processing Korean text.
Similarly, it follows the GPT-3 architecture.
\section{Supporting Material}\label{sec:support}
\begin{figure*}[h]
    \centering
    \includegraphics[width=\linewidth]{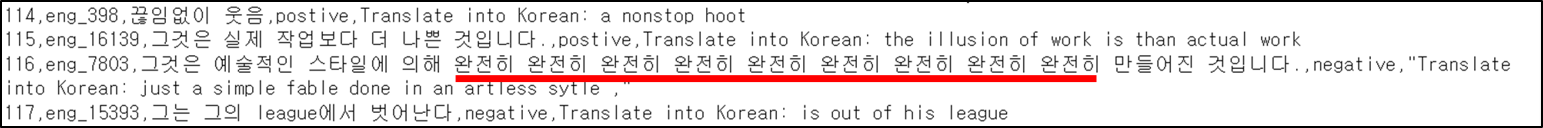} 
    \caption{Example of looping in translations using mT5. The word ``완전히(totaly or completely)'' was repeated countless times.
    }
\label{fig:looping}
\end{figure*}

Figure~\ref{fig:looping} shows mT5's malfunction `looping'.

\begin{figure}[h]
    \centering
    \includegraphics[width=\linewidth]{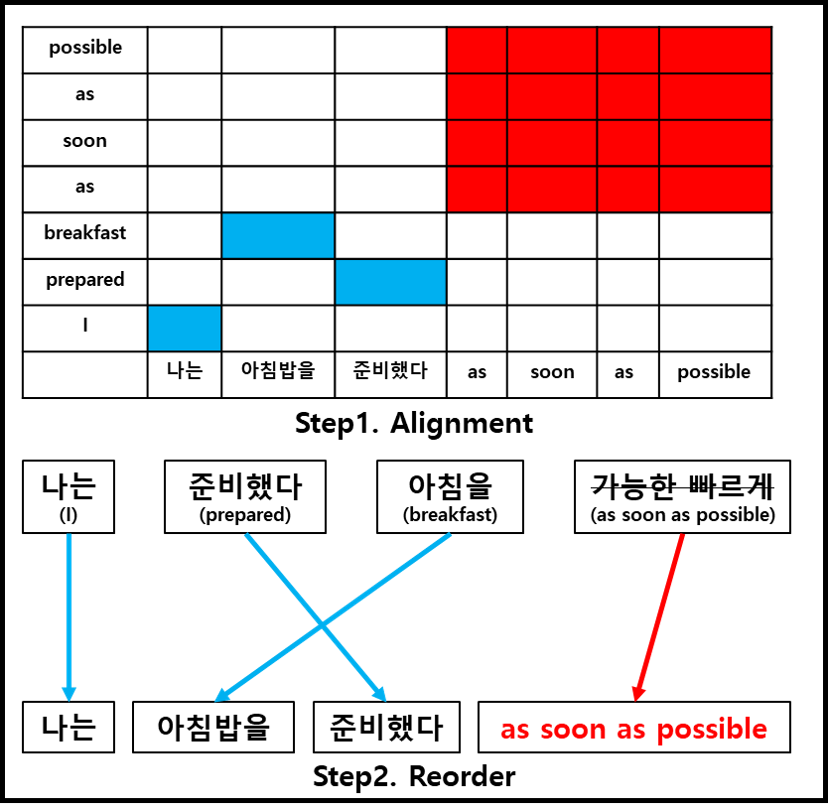} 
    \caption{Example of a mixed Korean-English sentence: When an input that does not exist in the learned probability table is encountered, the SMT outputs the expression in the source language as it is. The illustration shows that the expression ``as soon as possible'' is displayed in English due to the lack of information about this phrase.
    }
    \label{fig:mixedSMTexample}
\end{figure}

Figure~\ref{fig:mixedSMTexample} shows how SMT generate Korean-English mixed sentence.

\section{Implementation Details} \label{ap:impDetail}
\subsection{Development Environment}

\subsubsection{Hardware Specifications}
\begin{itemize}
    \item \textbf{GPU}: NVIDIA A100 40GB
\end{itemize}

\subsubsection{Software Environment}
\begin{itemize}
    \item \textbf{Operating System}: Ubuntu 20.04.5
    \item \textbf{Programming Language}: Python 3.8.10
\end{itemize}

\subsubsection{Key Libraries and Tools}
\begin{table}[h!]
\centering
\begin{tabular}{|l|l|}
\hline
\textbf{Library/Tool} & \textbf{Version} \\
\hline
PyTorch       & 2.1.2  \\
Torchvision   & 0.16.2 \\
Torchaudio    & 2.1.2  \\
Transformers  & 4.38.2 \\
Accelerate    & 0.27.2 \\
OpenAI        & 0.27.7 \\
\hline
\end{tabular}
\end{table}

\subsection{Instruction Sets}
Table~\ref{ap:instructions} shows prompt set for this study.
\begin{table*}[]
\begin{adjustbox}{width=\linewidth}
\begin{tabular}{|c|l|l|}
\hline
model                        & task        & prompt                                                                    \\ \hline
\multirow{4}{*}{MGPT/ko-GPT} & SA          & Sentiment Analysis:                                                       \\ \cline{2-3} 
                             & NLI         & Inference relation between premise and hypothesis:                        \\ \cline{2-3} 
                             & STS         & measure sementic similarity between sentence1 and sentence2 in   0.0-5.0: \\ \cline{2-3} 
                             & translation & Translate into Korean:                                                    \\ \hline
\multirow{4}{*}{mT5/pko-T5}  & SA          & Sentiment Analysis:                                                       \\ \cline{2-3} 
                             & NLI         & NLI:                                                                      \\ \cline{2-3} 
                             & STS         & STS:                                                                      \\ \cline{2-3} 
                             & translation & Translate into Korean:                                                    \\ \hline
\end{tabular}
\end{adjustbox}
\caption{Prompt Set for Experiments}\label{ap:instructions}
\end{table*}

\section{mT5 Experiment results}\label{ap:mt5Results}
Tables ~\ref{tab:samt5},  ~\ref{tab:nlimt5}, and  ~\ref{tab:stsmt5} show mT5's experiment results.
\begin{table*}[h]
\begin{adjustbox}{width=\linewidth}
\begin{tabular}{|l|l|l|l|l|l|l|l|l|l|l|l|l|l|l|l|l|l|l|}
\hline
             & 1k & 2k       & 3k       & 4k       & 5k       & 6k       & 7k       & 8k       & 9k       & 10k      & 12k      & 14k      & 16k      & 18k      & 20k      & 30k      & 40k      & 50k      \\ \hline
MT5\_EN      & 0  & 0.644045 & 0.644527 & 0.644746 & 0.643826 & 0.644286 & 0.642614 & 0.642614 & 0.645293 & 0.644195 & 0.650728 & 0.64363  & 0.647942 & 0.60883  & 0.601297 &          &          &          \\ \hline
MT5\_TR-GPT  & 0  & 0.404482 & 0.644068 & 0.644527 & 0.644527 & 0.644527 & 0.644527 & 0.644746 & 0.644527 & 0.646018 & 0.646111 & 0.644007 & 0.585694 & 0.575898 & 0.551376 &          &          &          \\ \hline
MT5\_PAG     & 0  & 0.632231 & 0.637    & 0.644527 & 0.644527 & 0.644527 & 0.645    & 0.644527 & 0.644527 & 0.644527 & 0.645    & 0.644527 & 0.643319 & 0.643255 & 0.658786 & 0.623162 & 0.640127 & 0.524161 \\ \hline
MT5\_PAG25   & 0  & 0.644022 & 0.645    & 0.644746 & 0.644527 & 0.644068 & 0.645    & 0.644068 & 0.643826 & 0.644527 & 0.645    & 0.644964 & 0.644527 & 0.645492 & 0.631311 & 0.596236 & 0.560427 & 0.656516 \\ \hline
MT5\_PAG50   & 0  & 0.644527 & 0.645    & 0.644527 & 0.644527 & 0.644527 & 0.645    & 0.644527 & 0.640763 & 0.644286 & 0.645    & 0.644964 & 0.643561 & 0.644527 & 0.642416 & 0.620624 & 0.66775  & 0.602317 \\ \hline
MT5\_PAG75   & 0  & 0.633793 & 0.628    & 0.625602 & 0.644527 & 0.644527 & 0.645    & 0.644527 & 0.644527 & 0.644746 & 0.645    & 0.644527 & 0.645798 & 0.644527 & 0.599804 & 0.647611 & 0.541966 & 0.66575  \\ \hline
MT5\_Moses   & 0  & 0.597122 & 0.628    & 0.643803 & 0.642445 & 0.644527 & 0.645    & 0.644746 & 0.644286 & 0.644527 & 0.645    & 0.644527 & 0.640057 & 0.647997 & 0.629205 & 0.631629 & 0.612291 & 0.590587 \\ \hline
MT5\_MGPT-TR & 0  & 0        & 0.644527 & 0.644527 & 0.644527 & 0.644527 & 0.644723 & 0.644527 & 0.644527 & 0.644527 & 0.643916 & 0.62851  & 0.648444 & 0.60623  & 0.64496  &          &          &          \\ \hline
MT5\_mT5-TR  & 0  & 0.001178 & 0.644527 & 0.643826 & 0.641766 & 0.644172 & 0.644527 & 0.644527 & 0.644527 & 0.644634 & 0.64492  & 0.636837 & 0.605634 & 0.636135 & 0.60039  &          &          &          \\ \hline
MT5\_KO      & 0  & 0.560894 & 0.642857 & 0.642246 & 0.642784 & 0.644068 & 0.643803 & 0.645691 & 0.644437 & 0.646797 & 0.651685 & 0.653119 & 0.67061  & 0.580201 & 0.702138 &          &          &          \\ \hline
\end{tabular}
\end{adjustbox}
\caption{SA experiment results using mT5. The model output a single label due to improper training.}
\label{tab:samt5}
\end{table*}
\begin{table*}[h]
\begin{adjustbox}{width=\linewidth}
\begin{tabular}{|l|l|l|l|l|l|l|l|l|l|l|l|l|l|l|l|l|l|l|}
\hline
             & 1k       & 2k       & 3k       & 4k       & 5k       & 6k       & 7k       & 8k       & 9k       & 10k      & 12k      & 14k      & 16k      & 18k      & 20k      & 30k      & 40k      & 50k      \\ \hline
MT5\_EN      & 0.332667 & 0.34     & 0.336    & 0.356    & 0.344    & 0.347333 & 0.346667 & 0.339333 & 0.337333 & 0.349333 & 0.351333 & 0.352    & 0.36     & 0.354667 & 0.339333 &          &          &          \\ \hline
MT5\_TR-GPT  & 0.322667 & 0.338    & 0.333333 & 0.341333 & 0.346667 & 0.345333 & 0.335333 & 0.340667 & 0.324    & 0.338667 & 0.332    & 0.334667 & 0.347333 & 0.34     & 0.392    &          &          &          \\ \hline
MT5\_PAG     & 0.202    & 0.309333 & 0.346    & 0.346    & 0.340667 & 0.34     & 0.34     & 0.338    & 0.371333 & 0.363    & 0.347333 & 0.350667 & 0.346667 & 0.342667 & 0.35     & 0.359333 & 0.368    & 0.397333 \\ \hline
MT5\_PAG25   & 0        & 0.341333 & 0.341333 & 0.353    & 0.366667 & 0.35     & 0.34     & 0.35     & 0.348667 & 0.331    & 0.330667 & 0.349333 & 0.335333 & 0.348    & 0.334    & 0.336    & 0.350667 & 0.356    \\ \hline
MT5\_PAG50   & 0.33     & 0.324    & 0.342    & 0.338    & 0.336667 & 0.337    & 0.346667 & 0.352667 & 0.337333 & 0.339    & 0.339333 & 0.323333 & 0.341333 & 0.334667 & 0.346    & 0.354667 & 0.36     & 0.319333 \\ \hline
MT5\_PAG75   & 0.344    & 0.326667 & 0.337333 & 0.34     & 0.345333 & 0.352    & 0.362667 & 0.318    & 0.336667 & 0.328    & 0.326    & 0.334    & 0.336    & 0.337333 & 0.316667 & 0.338667 & 0.337333 & 0.384    \\ \hline
MT5\_Moses   & 0.329333 & 0.339333 & 0.36     & 0.346    & 0.34     & 0.34     & 0.340667 & 0.337333 & 0.347333 & 0.348    & 0.347333 & 0.366    & 0.337333 & 0.333333 & 0.350667 & 0.336667 & 0.31     & 0.344667 \\ \hline
MT5\_MGPT-TR & 0        & 0.301333 & 0.336    & 0.339333 & 0.336    & 0.336667 & 0.344    & 0.344    & 0.344    & 0.348    & 0.336667 & 0.338    & 0.338667 & 0.342    & 0.341333 &          &          &          \\ \hline
MT5\_mT5-TR  & 0        & 0.328    & 0.338667 & 0.34     & 0.346    & 0.347333 & 0.340667 & 0.337333 & 0.342    & 0.34     & 0.339333 & 0.34     & 0.35     & 0.344667 & 0.336    &          &          &          \\ \hline
MT5\_KO      & 0.344    & 0.339333 & 0.35     & 0.348667 & 0.339333 & 0.336    & 0.344    & 0.338    & 0.355333 & 0.335333 & 0.345333 & 0.338    & 0.328667 & 0.338667 & 0.354667 &          &          &          \\ \hline
\end{tabular}
\end{adjustbox}
\caption{NLI experiment results using mT5. The model output a single label due to improper training.}
\label{tab:nlimt5}
\end{table*}
\begin{table*}[h]
\begin{adjustbox}{width=\linewidth}
\begin{tabular}{|l|l|l|l|l|l|l|l|}
\hline
             & 1k       & 2k       & 3k       & 4k       & 5k       & 6k       & 7k       \\ \hline
MT5\_EN      & -0.08886 & -0.11688 & 0.112826 & -0.08782 & -0.01175 & -0.05323 & -0.00847 \\ \hline
MT5\_TR-GPT  & -0.06812 & -0.03022 & -0.09011 & -0.163   & -0.10468 & -0.08874 & -0.07869 \\ \hline
MT5\_PAG     & -0.04207 & -0.162   & -0.06356 & 0.002619 & 0.035457 & 0.004099 & 0.00316  \\ \hline
MT5\_PAG25   & -0.14975 & -0.04355 & -0.06311 & -0.06307 & -0.06827 & -0.07657 & -0.08174 \\ \hline
MT5\_PAG50   & -0.19666 & -0.10183 & -0.01563 & -0.05113 & -0.06781 & -0.03932 & -0.06619 \\ \hline
MT5\_PAG75   & -0.04759 & -0.1402  & -0.09614 & -0.07074 & -0.07721 & -0.12111 & 0.053848 \\ \hline
MT5\_Moses   & -0.18494 & -0.09123 & -0.05138 & -0.00623 & 0.02206  & -0.06622 & 0.050311 \\ \hline
MT5\_MGPT-TR & -0.1     & -0.10514 & -0.08911 & -0.13977 & -0.02636 & -0.05928 & 0.038272 \\ \hline
MT5\_mT5-TR  & -0.1     & -0.14966 & -0.12913 & -0.18045 & 0.057898 & 0.003696 & 0.03223  \\ \hline
MT5\_KO      & -0.07757 & -0.02777 & -0.09689 & -0.02668 & -0.1     & 0.099926 & -0.1     \\ \hline
\end{tabular}
\end{adjustbox}
\caption{STS experiment results using mT5. The model output a single label due to improper training.}
\label{tab:stsmt5}
\end{table*}
\section{Detailed Experiment Results}\label{ap:detailedResults}
\subsection{SA}
Tables ~\ref{tab:samgpt}, ~\ref{tab:sakgpt}, ~\ref{tab:sakt5}, ~\ref{tab:nlimgpt}, ~\ref{tab:nlikGPT}, ~\ref{tab:nlikt5},~\ref{tab:STSmgpt}, ~\ref{tab:stsKGPT}, and ~\ref{tab:stsKT5} shows full result of our experiments.

\begin{table*}[]
\begin{adjustbox}{width=\linewidth}
\begin{tabular}{|l|l|l|l|l|l|l|l|l|l|l|l|l|l|l|l|l|l|l|}
\hline
              & 1k    & 2k    & 3k    & 4k    & 5k    & 6k    & 7k    & 8k    & 9k    & 10k   & 12k   & 14k   & 16k   & 18k   & 20k   & 30k   & 40k   & 50k   \\ \hline
MGPT\_EN      & 0.673 & 0.688 & 0.708 & 0.696 & 0.716 & 0.712 & 0.7   & 0.716 & 0.722 & 0.717 & 0.73  & 0.716 & 0.729 & 0.725 & 0.727 &       &       &       \\ \hline
MGPT\_TR-GPT  & 0.723 & 0.741 & 0.745 & 0.748 & 0.748 & 0.758 & 0.758 & 0.755 & 0.755 & 0.752 & 0.765 & 0.765 & 0.769 & 0.766 & 0.772 &       &       &       \\ \hline
MGPT\_PAD     & 0.696 & 0.721 & 0.731 & 0.735 & 0.732 & 0.732 & 0.735 & 0.737 & 0.742 & 0.754 & 0.752 & 0.752 & 0.762 & 0.759 & 0.77  & 0.759 & 0.764 & 0.77  \\ \hline
MGPT\_PAD25   & 0.674 & 0.699 & 0.703 & 0.707 & 0.717 & 0.67  & 0.706 & 0.725 & 0.722 & 0.724 & 0.728 & 0.731 & 0.734 & 0.727 & 0.733 & 0.729 & 0.745 & 0.738 \\ \hline
MGPT\_PAD50   & 0.697 & 0.689 & 0.711 & 0.715 & 0.722 & 0.725 & 0.719 & 0.724 & 0.735 & 0.716 & 0.725 & 0.738 & 0.736 & 0.739 & 0.742 & 0.734 & 0.739 & 0.738 \\ \hline
MGPT\_PAD75   & 0.721 & 0.709 & 0.704 & 0.704 & 0.726 & 0.732 & 0.727 & 0.728 & 0.729 & 0.724 & 0.734 & 0.745 & 0.732 & 0.735 & 0.729 & 0.745 & 0.741 & 0.742 \\ \hline
MGPT\_Moses   & 0.728 & 0.735 & 0.736 & 0.735 & 0.732 & 0.736 & 0.743 & 0.75  & 0.747 & 0.759 & 0.749 & 0.748 & 0.753 & 0.77  & 0.742 & 0.765 & 0.759 & 0.755 \\ \hline
MGPT\_MGPT-TR & 0.643 & 0.667 & 0.639 & 0.705 & 0.714 & 0.749 & 0.719 & 0.725 & 0.742 & 0.734 & 0.755 & 0.751 & 0.75  & 0.732 & 0.756 &       &       &       \\ \hline
MGPT\_mT5-TR  & 0.634 & 0.656 & 0.651 & 0.719 & 0.722 & 0.739 & 0.735 & 0.728 & 0.73  & 0.714 & 0.736 & 0.741 & 0.747 & 0.758 & 0.757 &       &       &       \\ \hline
MGPT\_KO      & 0.728 & 0.755 & 0.758 & 0.762 & 0.777 & 0.769 & 0.792 & 0.822 & 0.803 & 0.821 & 0.83  & 0.821 & 0.817 & 0.821 & 0.826 &       &       &       \\ \hline
\end{tabular}
\end{adjustbox}
\caption{SA experiment results using MGPT}
\label{tab:samgpt}
\end{table*}
\begin{table*}[h]
\begin{adjustbox}{width=\linewidth}
\begin{tabular}{|l|l|l|l|l|l|l|l|l|l|l|l|l|l|l|l|l|l|l|}
\hline
              & 1k    & 2k    & 3k    & 4k    & 5k    & 6k    & 7k    & 8k    & 9k    & 10k   & 12k   & 14k   & 16k   & 18k   & 20k   & 30k   & 40k   & 50k   \\ \hline
KGPT\_TR-GPT  & 0.834 & 0.832 & 0.835 & 0.839 & 0.845 & 0.849 & 0.854 & 0.851 & 0.857 & 0.857 & 0.859 & 0.867 & 0.863 & 0.868 & 0.866 &       &       &       \\ \hline
KGPT\_PAD     & 0.824 & 0.819 & 0.826 & 0.832 & 0.833 & 0.838 & 0.835 & 0.833 & 0.839 & 0.843 & 0.845 & 0.847 & 0.845 & 0.852 & 0.849 & 0.86  & 0.853 & 0.862 \\ \hline
KGPT\_Moses   & 0.823 & 0.819 & 0.823 & 0.827 & 0.83  & 0.816 & 0.822 & 0.843 & 0.849 & 0.834 & 0.845 & 0.849 & 0.855 & 0.864 & 0.856 & 0.855 & 0.859 & 0.863 \\ \hline
KGPT\_MGPT-TR & 0.766 & 0.79  & 0.791 & 0.809 & 0.802 & 0.82  & 0.8   & 0.816 & 0.826 & 0.837 & 0.814 & 0.813 & 0.831 & 0.838 & 0.847 &       &       &       \\ \hline
KGPT\_mT5-TR  & 0.816 & 0.819 & 0.819 & 0.818 & 0.826 & 0.832 & 0.827 & 0.823 & 0.831 & 0.832 & 0.841 & 0.847 & 0.835 & 0.853 & 0.86  &       &       &       \\ \hline
KGPT\_KO      & 0.851 & 0.876 & 0.874 & 0.871 & 0.875 & 0.872 & 0.881 & 0.893 & 0.892 & 0.891 & 0.892 & 0.9   & 0.895 & 0.895 & 0.902 &       &       &       \\ \hline
\end{tabular}
\end{adjustbox}
\caption{SA experiment results using koGPT}
\label{tab:sakgpt}
\end{table*}
\begin{table*}[h]
\begin{adjustbox}{width=\linewidth}
\begin{tabular}{|l|l|l|l|l|l|l|l|l|l|l|l|l|l|l|l|l|l|l|}
\hline
              & 1k    & 2k    & 3k    & 4k    & 5k    & 6k    & 7k    & 8k    & 9k    & 10k   & 12k   & 14k   & 16k   & 18k   & 20k   & 30k   & 40k   & 50k   \\ \hline
KmT5\_TR-GPT  & 0.723 & 0.793 & 0.817 & 0.827 & 0.828 & 0.828 & 0.842 & 0.842 & 0.852 & 0.848 & 0.849 & 0.855 & 0.855 & 0.86  & 0.863 &       &       &       \\ \hline
KmT5\_PAD     & 0.747 & 0.778 & 0.784 & 0.795 & 0.804 & 0.816 & 0.819 & 0.821 & 0.832 & 0.828 & 0.83  & 0.831 & 0.832 & 0.842 & 0.841 & 0.843 & 0.846 & 0.845 \\ \hline
KmT5\_Moses   & 0.77  & 0.779 & 0.786 & 0.804 & 0.798 & 0.822 & 0.824 & 0.827 & 0.82  & 0.819 & 0.829 & 0.832 & 0.831 & 0.838 & 0.839 & 0.846 & 0.846 & 0.852 \\ \hline
KmT5\_MGPT-TR & 0.642 & 0.727 & 0.784 & 0.777 & 0.797 & 0.803 & 0.814 & 0.811 & 0.814 & 0.826 & 0.824 & 0.814 & 0.827 & 0.834 & 0.839 &       &       &       \\ \hline
KmT5\_mT5-TR  & 0.667 & 0.605 & 0.799 & 0.78  & 0.813 & 0.824 & 0.828 & 0.829 & 0.839 & 0.836 & 0.845 & 0.845 & 0.852 & 0.851 & 0.852 &       &       &       \\ \hline
KmT5\_KO      & 0.816 & 0.832 & 0.838 & 0.856 & 0.863 & 0.861 & 0.873 & 0.879 & 0.881 & 0.885 & 0.885 & 0.892 & 0.895 & 0.907 & 0.902 &       &       &       \\ \hline
\end{tabular}

\end{adjustbox}
\caption{SA experiment results using kT5}
\label{tab:sakt5}
\end{table*}
\begin{table*}[h]
\begin{adjustbox}{width=\linewidth}
\begin{tabular}{|l|l|l|l|l|l|l|l|l|l|l|l|l|l|l|l|l|l|l|}
\hline
              & 1k    & 2k    & 3k    & 4k    & 5k    & 6k    & 7k    & 8k    & 9k    & 10k   & 12k   & 14k   & 16k   & 18k   & 20k   & 30k   & 40k   & 50k   \\ \hline
MGPT\_EN      & 0.25  & 0.237 & 0.265 & 0.269 & 0.282 & 0.3   & 0.357 & 0.307 & 0.433 & 0.421 & 0.417 & 0.473 & 0.474 & 0.589 & 0.572 &       &       &       \\ \hline
MGPT\_TR-GPT  & 0.285 & 0.319 & 0.353 & 0.389 & 0.386 & 0.507 & 0.515 & 0.569 & 0.544 & 0.629 & 0.672 & 0.657 & 0.666 & 0.722 & 0.711 &       &       &       \\ \hline
MGPT\_PAD     & 0.315 & 0.323 & 0.345 & 0.352 & 0.369 & 0.392 & 0.385 & 0.509 & 0.494 & 0.51  & 0.639 & 0.514 & 0.665 & 0.671 & 0.681 & 0.687 & 0.693 & 0.701 \\ \hline
MGPT\_PAD25   & 0.287 & 0.297 & 0.323 & 0.328 & 0.34  & 0.375 & 0.36  & 0.386 & 0.427 & 0.497 & 0.46  & 0.521 & 0.626 & 0.641 & 0.649 & 0.659 & 0.667 & 0.675 \\ \hline
MGPT\_PAD50   & 0.289 & 0.303 & 0.335 & 0.353 & 0.363 & 0.385 & 0.373 & 0.412 & 0.421 & 0.433 & 0.491 & 0.444 & 0.659 & 0.667 & 0.672 & 0.676 & 0.687 & 0.696 \\ \hline
MGPT\_PAD75   & 0.317 & 0.341 & 0.346 & 0.362 & 0.369 & 0.41  & 0.375 & 0.416 & 0.437 & 0.448 & 0.566 & 0.478 & 0.53  & 0.627 & 0.669 & 0.671 & 0.684 & 0.687 \\ \hline
MGPT\_Moses   & 0.312 & 0.321 & 0.343 & 0.36  & 0.367 & 0.439 & 0.382 & 0.495 & 0.505 & 0.515 & 0.555 & 0.531 & 0.612 & 0.635 & 0.663 & 0.671 & 0.674 & 0.682 \\ \hline
MGPT\_MGPT-TR & 0.25  & 0.282 & 0.295 & 0.329 & 0.346 & 0.369 & 0.377 & 0.41  & 0.425 & 0.446 & 0.514 & 0.469 & 0.578 & 0.601 & 0.61  &       &       &       \\ \hline
MGPT\_mT5-TR  & 0.337 & 0.369 & 0.393 & 0.407 & 0.441 & 0.476 & 0.455 & 0.497 & 0.551 & 0.557 & 0.593 & 0.579 & 0.67  & 0.673 & 0.679 &       &       &       \\ \hline
MGPT\_KO      & 0.309 & 0.332 & 0.357 & 0.384 & 0.515 & 0.507 & 0.583 & 0.565 & 0.573 & 0.673 & 0.686 & 0.692 & 0.667 & 0.745 & 0.724 &       &       &       \\ \hline
\end{tabular}
\end{adjustbox}
\caption{NLI experiment results using MGPT}
\label{tab:nlimgpt}
\end{table*}
\begin{table*}[h]
\begin{adjustbox}{width=\linewidth}
\begin{tabular}{|l|l|l|l|l|l|l|l|l|l|l|l|l|l|l|l|l|l|l|}
\hline
              & 1k    & 2k    & 3k    & 4k    & 5k    & 6k    & 7k    & 8k    & 9k    & 10k   & 12k   & 14k   & 16k   & 18k   & 20k   & 30k   & 40k   & 50k   \\ \hline
KGPT\_TR-GPT  & 0.691 & 0.73  & 0.733 & 0.731 & 0.765 & 0.748 & 0.754 & 0.761 & 0.77  & 0.789 & 0.773 & 0.78  & 0.766 & 0.773 & 0.782 &       &       &       \\ \hline
KGPT\_PAD     & 0.663 & 0.696 & 0.708 & 0.715 & 0.72  & 0.739 & 0.746 & 0.742 & 0.749 & 0.751 & 0.746 & 0.747 & 0.762 & 0.767 & 0.754 & 0.763 & 0.754 & 0.767 \\ \hline
KGPT\_Moses   & 0.689 & 0.701 & 0.716 & 0.723 & 0.728 & 0.733 & 0.735 & 0.755 & 0.757 & 0.74  & 0.764 & 0.767 & 0.757 & 0.769 & 0.76  & 0.747 & 0.771 & 0.766 \\ \hline
KGPT\_MGPT-TR & 0.513 & 0.601 & 0.621 & 0.663 & 0.663 & 0.691 & 0.669 & 0.715 & 0.701 & 0.699 & 0.747 & 0.727 & 0.738 & 0.729 & 0.758 &       &       &       \\ \hline
KGPT\_mT5-TR  & 0.555 & 0.639 & 0.672 & 0.669 & 0.689 & 0.708 & 0.695 & 0.707 & 0.733 & 0.73  & 0.747 & 0.731 & 0.75  & 0.758 & 0.753 &       &       &       \\ \hline
KGPT\_KO      & 0.723 & 0.74  & 0.767 & 0.777 & 0.784 & 0.78  & 0.791 & 0.779 & 0.794 & 0.802 & 0.811 & 0.801 & 0.816 & 0.832 & 0.834 &       &       &       \\ \hline
\end{tabular}
\end{adjustbox}
\caption{NLI experiment results using koGPT}
\label{tab:nlikGPT}
\end{table*}
\begin{table*}[h]
\begin{adjustbox}{width=\linewidth}
\begin{tabular}{|l|l|l|l|l|l|l|l|l|l|l|l|l|l|l|l|l|l|l|}
\hline
              & 1k    & 2k    & 3k    & 4k    & 5k    & 6k    & 7k    & 8k    & 9k    & 10k   & 12k   & 14k   & 16k   & 18k   & 20k   & 30k   & 40k   & 50k   \\ \hline
KmT5\_TR-GPT  & 0.45  & 0.6   & 0.601 & 0.688 & 0.687 & 0.713 & 0.689 & 0.701 & 0.7   & 0.721 & 0.731 & 0.751 & 0.754 & 0.764 & 0.75  &       &       &       \\ \hline
KmT5\_PAD     & 0.38  & 0.505 & 0.579 & 0.597 & 0.618 & 0.687 & 0.687 & 0.706 & 0.712 & 0.714 & 0.726 & 0.731 & 0.728 & 0.732 & 0.732 & 0.739 & 0.741 & 0.741 \\ \hline
KmT5\_Moses   & 0.437 & 0.551 & 0.569 & 0.6   & 0.666 & 0.677 & 0.689 & 0.711 & 0.703 & 0.685 & 0.732 & 0.709 & 0.73  & 0.729 & 0.726 & 0.733 & 0.747 & 0.741 \\ \hline
KmT5\_MGPT-TR & 0.433 & 0.44  & 0.508 & 0.579 & 0.605 & 0.635 & 0.621 & 0.657 & 0.663 & 0.675 & 0.682 & 0.703 & 0.711 & 0.717 & 0.723 &       &       &       \\ \hline
KmT5\_mT5-TR  & 0.359 & 0.439 & 0.497 & 0.579 & 0.613 & 0.65  & 0.667 & 0.695 & 0.695 & 0.719 & 0.719 & 0.741 & 0.746 & 0.743 & 0.734 &       &       &       \\ \hline
KmT5\_KO      & 0.513 & 0.649 & 0.64  & 0.708 & 0.665 & 0.723 & 0.727 & 0.732 & 0.765 & 0.752 & 0.779 & 0.772 & 0.794 & 0.805 & 0.803 &       &       &       \\ \hline
\end{tabular}
\end{adjustbox}
\caption{NLI experiment results using kT5}
\label{tab:nlikt5}
\end{table*}
\begin{table*}[h]
\begin{adjustbox}{width=\linewidth}
\begin{tabular}{|l|l|l|l|l|l|l|l|}
\hline
              & 1k    & 2k    & 3k    & 4k    & 5k    & 6k    & 7k    \\ \hline
MGPT\_EN      & 0.553 & 0.513 & 0.694 & 0.651 & 0.665 & 0.655 & 0.819 \\ \hline
MGPT\_TR-GPT  & 0.574 & 0.547 & 0.752 & 0.711 & 0.691 & 0.691 & 0.851 \\ \hline
MGPT\_PAD     & 0.55  & 0.553 & 0.735 & 0.693 & 0.677 & 0.684 & 0.844 \\ \hline
MGPT\_PAD25   & 0.545 & 0.559 & 0.714 & 0.67  & 0.682 & 0.676 & 0.82  \\ \hline
MGPT\_PAD50   & 0.548 & 0.535 & 0.731 & 0.668 & 0.678 & 0.671 & 0.838 \\ \hline
MGPT\_PAD75   & 0.525 & 0.546 & 0.743 & 0.675 & 0.667 & 0.668 & 0.813 \\ \hline
MGPT\_Moses   & 0.523 & 0.575 & 0.736 & 0.659 & 0.671 & 0.682 & 0.821 \\ \hline
MGPT\_MGPT-TR & 0.53  & 0.59  & 0.745 & 0.688 & 0.671 & 0.692 & 0.841 \\ \hline
MGPT\_mT5-TR  & 0.529 & 0.58  & 0.762 & 0.698 & 0.704 & 0.685 & 0.84  \\ \hline
MGPT\_KO      & 0.548 & 0.64  & 0.779 & 0.748 & 0.744 & 0.745 & 0.88  \\ \hline
\end{tabular}
\end{adjustbox}
\caption{STS experiment results using MGPT}
\label{tab:STSmgpt}
\end{table*}
\begin{table*}[h]
\begin{adjustbox}{width=\linewidth}
\begin{tabular}{|l|l|l|l|l|l|l|l|}
\hline
              & 1k    & 2k    & 3k    & 4k    & 5k    & 6k    & 7k    \\ \hline
KGPT\_TR-GPT  & 0.811 & 0.818 & 0.824 & 0.828 & 0.83  & 0.833 & 0.831 \\ \hline
KGPT\_PAD     & 0.803 & 0.814 & 0.818 & 0.821 & 0.826 & 0.824 & 0.829 \\ \hline
KGPT\_Moses   & 0.804 & 0.818 & 0.816 & 0.824 & 0.828 & 0.83  & 0.824 \\ \hline
KGPT\_MGPT-TR & 0.53  & 0.777 & 0.776 & 0.772 & 0.789 & 0.798 & 0.819 \\ \hline
KGPT\_mT5-TR  & 0.776 & 0.799 & 0.815 & 0.817 & 0.811 & 0.821 & 0.816 \\ \hline
KGPT\_KO      & 0.819 & 0.827 & 0.834 & 0.838 & 0.842 & 0.843 & 0.844 \\ \hline
\end{tabular}

\end{adjustbox}
\caption{STS experiment results using koGPT}
\label{tab:stsKGPT}
\end{table*}
\begin{table*}[h]
\begin{adjustbox}{width=\linewidth}
\begin{tabular}{|l|l|l|l|l|l|l|l|}
\hline
              & 1k    & 2k    & 3k    & 4k    & 5k    & 6k    & 7k    \\ \hline
KmT5\_TR-GPT  & 0.834 & 0.853 & 0.853 & 0.853 & 0.858 & 0.856 & 0.863 \\ \hline
KmT5\_PAD     & 0.822 & 0.831 & 0.835 & 0.84  & 0.854 & 0.855 & 0.855 \\ \hline
KmT5\_Moses   & 0.834 & 0.841 & 0.843 & 0.834 & 0.851 & 0.853 & 0.852 \\ \hline
KmT5\_MGPT-TR & 0.045 & 0.769 & 0.754 & 0.822 & 0.811 & 0.852 & 0.844 \\ \hline
KmT5\_mT5-TR  & 0.352 & 0.762 & 0.795 & 0.815 & 0.836 & 0.827 & 0.817 \\ \hline
KmT5\_KO      & 0.848 & 0.862 & 0.868 & 0.869 & 0.873 & 0.873 & 0.875 \\ \hline
\end{tabular}

\end{adjustbox}
\caption{STS experiment results using kT5}
\label{tab:stsKT5}
\end{table*}

\end{document}